\documentclass{article}
\usepackage{arxiv}

\usepackage[utf8]{inputenc} 
\usepackage[T1]{fontenc}    
\usepackage{hyperref}       
\usepackage{url}            
\usepackage{booktabs}       
\usepackage{amsfonts}       
\usepackage{nicefrac}       
\usepackage{microtype}      
\usepackage{graphicx}
\graphicspath{ {./images/} }
\usepackage{makeidx}         
\usepackage{multicol}        
\usepackage[bottom]{footmisc}
\usepackage{newtxmath}       
\usepackage{subcaption}
\usepackage[export]{adjustbox}

\newcommand{\eg}{\textit{e.g.}, }

\title{TrashCan: A Semantically-Segmented Dataset towards Visual Detection of Marine Debris}

\author{Jungseok Hong$^{1}$, Michael Fulton$^{2}$, and Junaed Sattar$^{3}$
\thanks{The authors are with the Department of Computer Science and Engineering, Minnesota Robotics Institute, University of Minnesota--Twin Cities, 100 Union St SE, Minneapolis, MN, 55455, USA
{\tt\small \{$^{1}$jungseok, $^{2}$fulto081, $^{3}$junaed\} at umn.edu.}}
}

\begin{document}
\maketitle
\section{Motivation, Problem Statement, and Related Work}
\label{sec:introduction}
Marine debris poses a significant threat to the aquatic ecosystem and has been an ever-increasing challenge to tackle. 
Different environmental and government agencies have proposed and adopted a number of approaches for cleanup of underwater trash, though few have been broadly successful. 
Moreover, the difficulty of the cleanup task increases many folds because of the need to precisely detect trash deposits and locate them, and subsequently remove such trash while protecting underwater flora and fauna. 
Vision-equipped autonomous underwater vehicles (AUVs) could be used to address these challenges. 
Along with underwater visual localization algorithms~\cite{Mo2019IROS}, a robust trash detection feature will provide AUVs with the capability to both detect and locate trash, and possibly remove it themselves or make it easier for manned cleanup missions. 
A precise \textit{underwater} trash detector, however, would need a significant dataset to train deep neural nets (\eg CNNs) for visually detecting debris which are often deformed and take non-rigid shapes.
Unfortunately, no such datasets have been made available to the public.

This paper presents \textit{TrashCan}, a large dataset comprised of images of underwater trash collected from a variety of sources~\cite{JAMSTECDebri}, annotated both using bounding boxes and segmentation labels, for development of robust detectors of marine debris.
The dataset has two versions, TrashCan-Material and TrashCan-Instance, corresponding to different object class configurations.
The eventual goal is to develop efficient and accurate trash detection methods suitable for onboard robot deployment. Along with information about the construction and sourcing of the TrashCan dataset, we present initial results of instance segmentation from Mask R-CNN~\cite{he2017mask} and object detection from Faster R-CNN~\cite{ren2015faster}.
These do not represent the best possible detection results but provides an initial baseline for future work in instance segmentation and object detection on the TrashCan dataset.

 While the topic of removing marine debris from surface and subsea environments has been a topic of scientific inquiry for some time, autonomous detection of trash has been less studied. 
Researchers have studied at the removal of debris from the ocean surface ~\cite{dianna.parker_detecting_2015}, mapping trash on beaches using LIDAR~\cite{ge_semi-automatic_2016}, and using forward-looking sonar (FLS) imagery to detect underwater debris by training a deep convolutional neural network (CNN)~\cite{Toro2016Trash}.
More recently, we presented an initial evaluation of state-of-the-art object detection algorithms for trash detection, using a predecessor of TrashCan~\cite{Fulton2019ICRA_Trash}, and showed how such datasets could be effectively improved with data produced by generative models~\cite{Hong2020ICRA}. 
TrashCan is a next step into autonomous detection and removal of trash: a larger, more detailed, and more varied dataset of marine debris for training deep neural networks for object detection. 

\section{The TrashCan Dataset}
\label{sec:dataset}

\begin{figure}[t!]
    \centering
    \begin{subfigure}[b]{0.7\linewidth}
        \centering
        \includegraphics[width=1\textwidth]{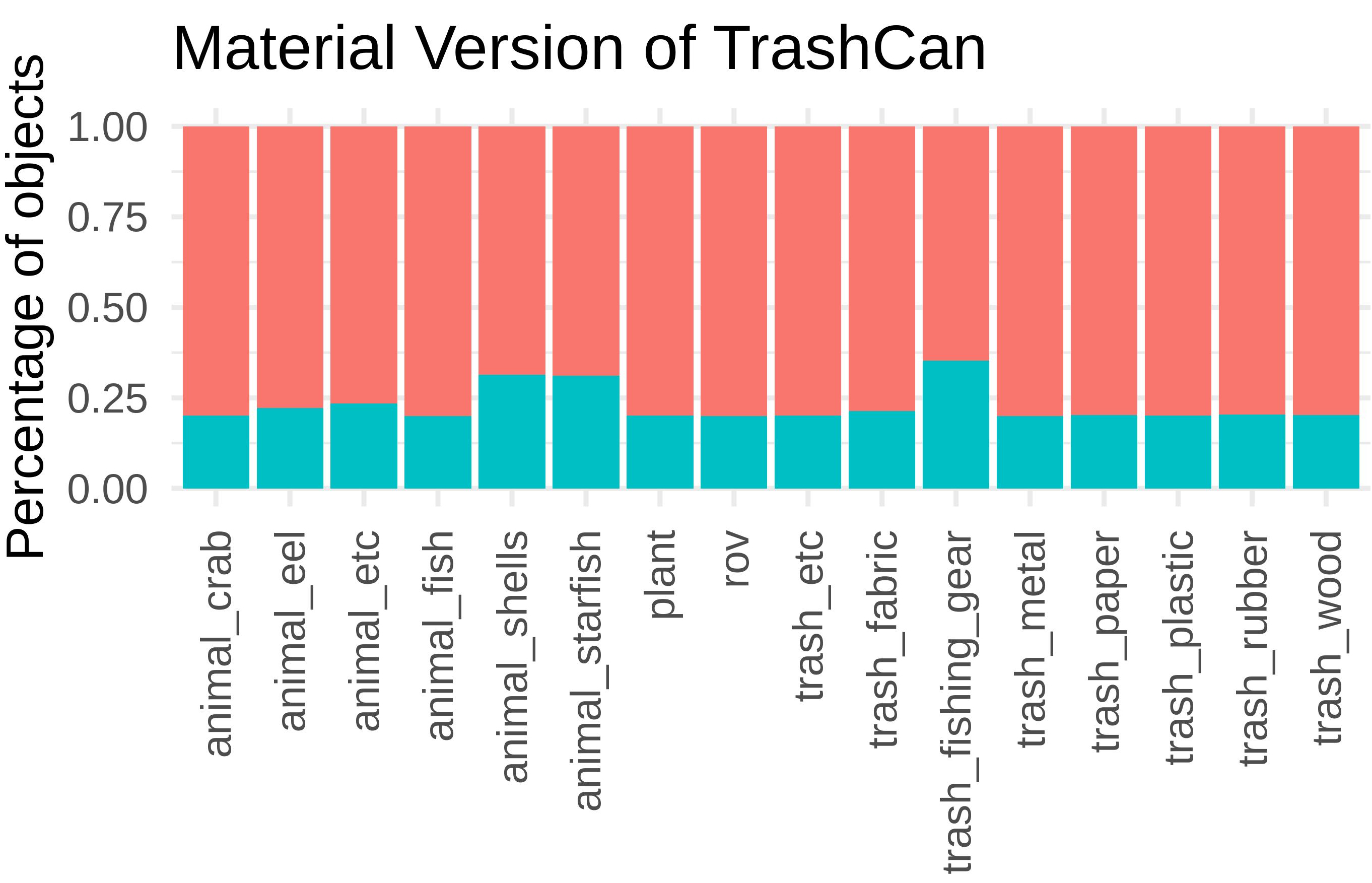}
        \caption{TrashCan-Material}
    \end{subfigure}%
    \hfill
    \begin{subfigure}[b]{0.7\linewidth}
        \centering
        \includegraphics[width=1\textwidth]{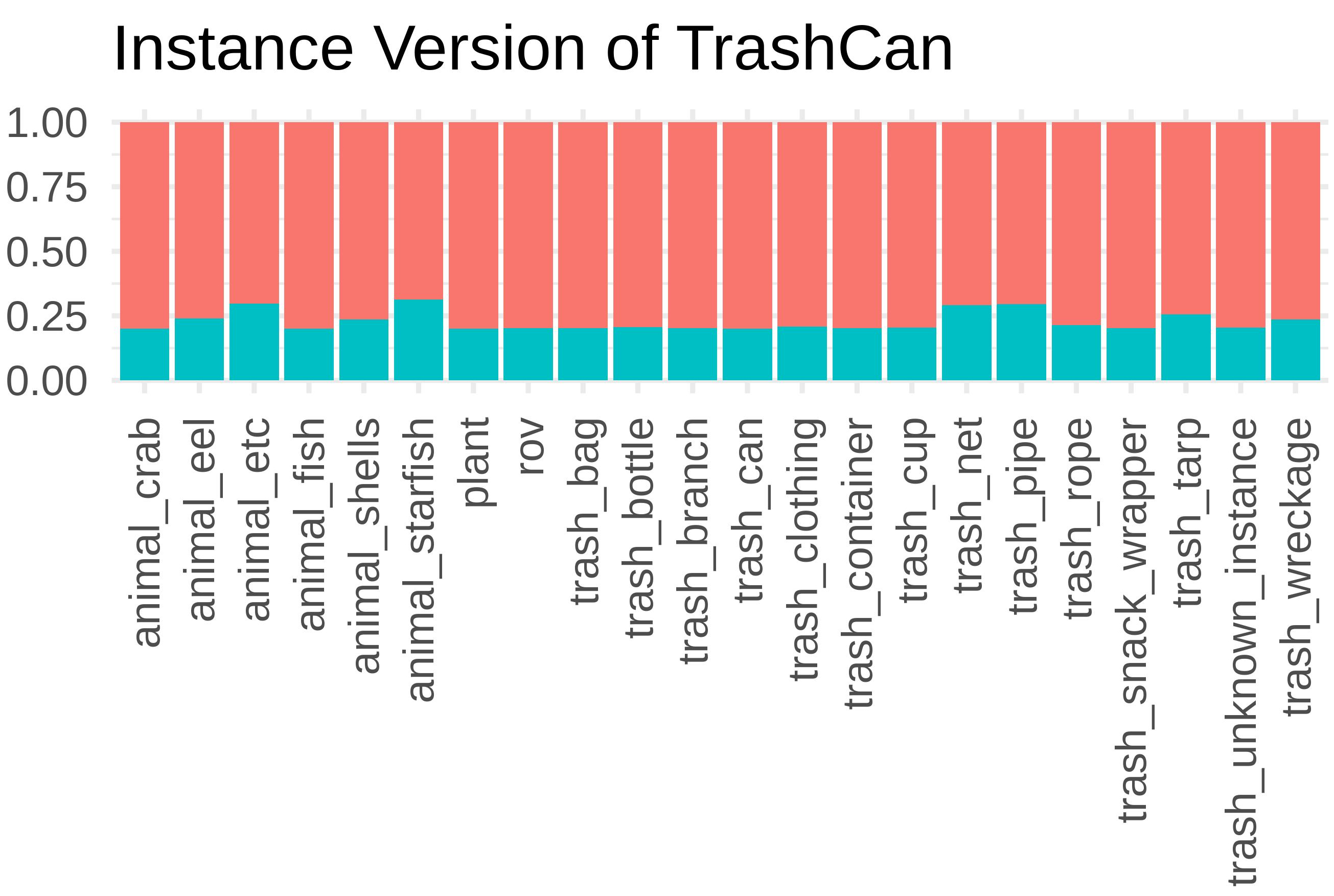}
        \caption{TrashCan-Instance}
    \end{subfigure}%
    \caption{Data split between training (pink) and validation (blue) sets per object for the two versions of the dataset.}
    \label{fig:dataset_split}
    \vspace{10mm}
\end{figure}
The TrashCan dataset is comprised of annotated images ($7,212$ images currently) which contain observations of trash, ROVs, and a wide variety of undersea flora and fauna. 
The annotations in this dataset take the format of instance segmentation annotations: bitmaps containing a mask marking which pixels in the image contain each object. 
While datasets have previously been created containing bounding box level annotations of trash in marine environments, including one of our own creation~\cite{Fulton2019ICRA_Trash}, TrashCan is, to the best of our knowledge, the first instance-segmentation annotated dataset of underwater trash. 

\textbf{Dataset Source and Composition:}
The imagery in TrashCan is sourced from the J-EDI  (JAMSTEC  E-Library of  Deep-sea  Images) dataset~\cite{JAMSTECDebri}, curated by the Japan Agency of Marine Earth Science and Technology (JAMSTEC). 
This dataset contains videos from ROVs operated by JAMSTEC since 1982, largely in the sea of Japan.
A small portion of these videos contain observations of marine debris, and it is from these that all of our trash data is sourced, nearly one thousand videos of varying lengths. 
In addition to the videos of marine debris, additional videos were selected to diversify the biological objects present in the dataset.

\textbf{Annotation Process and Tools:}
Once the videos had been selected, they were preprocessed by extracting frames from each video at a rate of one frame per second to create a large folder of frames for each video. 
Once this was done, the videos were combined into similarly-sized portions and uploaded to Supervisely~\cite{Supervisely}, an online image annotation tool. 
Once uploaded, the images were annotated by a team of 21 people, one image at a time.
This took approximately $1,500$ work hours, over the course of several months.
if an image was considered acceptable for labeling, the person to whom it was assigned drew a segmentation mask over it, marking it as one of four classes: \textit{trash} (any marine debris), \textit{rov} (any man-made item intentionally placed in the scene), \textit{bio} (plants and animals), and \textit{unknown} (used to mark unknown objects). 
Trash objects were additionally tagged by material (\eg metal, plastic), instance (\eg cup, bag, container), along with binary tags indicating overgrowth, significant decay, or crushed/broken items. 
Bio objects were tagged as either plant or animal, and in the case of animals, given a tag with the type (\eg crab, fish, eel). 
ROV and unknown class objects required no additional tags.


\textbf{Object Class Versions:}
To prepare the dataset for use in training deep networks, it was converted from a custom JSON format to the COCO format~\cite{lin2014microsoft}. 
This involved converting the bitmap masks into COCO annotations comprised of polygon vertices, and transforming the classes and tags of Supervisely annotations into classes for COCO objects. 
We converted all objects into one of two dataset versions: TrashCan-Material and TrashCan-Instance, so named for the tag data used to differentiate between different types of trash. 
In the material version, every trash object was given a class name following the pattern \textit{trash\_[material\_name]} (\eg \textit{trash\_paper}, \textit{trash\_plastic}), as long as the given material had more than 50 objects in the dataset. 
Those with fewer examples were given the class \textit{trash\_etc}, the same class used by annotators when the material of the object was unknown. 
Similarly, for the TrashCan-Instance version, trash classes were generated using instance tags which approximated the type of object that was being annotated (\eg \textit{trash\_cup}, \textit{trash\_bag}).  
The same cutoff of 50 objects was used, with the catch-all class being \textit{trash\_unknown\_instance}.
In both versions, any object labeled with the unknown class was typically added to the catch-all trash class.
The ROV class remained singular for both versions, while bio objects were either turned into \textit{plant} or \textit{animal\_[animal\_type]} (\eg \textit{animal\_starfish}, \textit{animal\_crab}) classes, based on tags applied to the object.

\section{Baseline Experiments}
\label{sec:experiments}
We present experiments with state-of-the-art instance segmentation and object detection models using two example datasets to provide a baseline for future model development. 
For the following experiments, we use the Pytorch Detectron2~\cite{wu2019detectron2} library and metrics introduced in COCO dataset to establish a baseline evaluation.

\textbf{Detection Experiments:}
\label{sec:experiments_detection}
For object detection, we employed Faster R-CNN with a ResNeXt-101-FPN (X-101-FPN)~\cite{Xie2016} backbone. 
The model was trained on a pre-trained model with COCO dataset with an Nvidia Titan XP for $10,000$ iterations.

\textbf{Segmentation Experiments:}
\label{sec:experiments_sgementation}
Mask R-CNN with X-101-FPN was chosen for instance segmentation task. 
The model was initialized with weights from the COCO dataset and trained for $15,000$ iterations using an Nvidia Titan V.
\begin{table}[h!]
    \caption{Overall metrics for each combination of dataset and model.}
    \vspace{3mm}
    \centering
    \begin{tabular}{p{2cm}p{2cm}p{1cm}p{1cm}p{1cm}p{1cm}p{1cm}p{1cm}}
    \hline
        Method & Dataset & AP & AP$_{50}$ & AP$_{75}$ & AP$_{S}$ & AP$_{M}$ & AP$_{L}$ \\\hline \hline
         Mask R-CNN & Instance & 30.0 &	55.3 & 29.4 & 23.2 & 31.7 &	48.6  \\ 
         Mask R-CNN & Material & 28.2 &	54.0 & 25.6 & 24.1	& 28.7 & 41.8 \\ \hline
         Faster R-CNN & Instance & 34.5 & 55.4 & 38.1 &	27.6 &	36.2 &	51.4 \\
         Faster R-CNN & Material & 29.1 & 51.2 & 27.8 &	28.2 & 30.2 & 40.0 \\ \hline
    \end{tabular}
    \label{tab:detection_instance_results}
    \vspace{10mm}
\end{table}
\section{Results and Conclusion}
\label{sec:Results}
\begin{figure*}[t!]
\vspace{35mm}
\setlength{\lineskip}{0pt}
\centering
\setlength\tabcolsep{1.pt}
\renewcommand{\arraystretch}{0.5}
  \begin{tabular}{c@{\extracolsep{0.1cm}}c@{\extracolsep{0.1cm}}c@{\extracolsep{0.1cm}}c@{\extracolsep{0.1cm}}c}
  \raisebox{4.5\normalbaselineskip}[0pt][0pt]{\rotatebox[origin=c]{90}{Seg-Instance}} &
    \includegraphics[width=.315\textwidth]{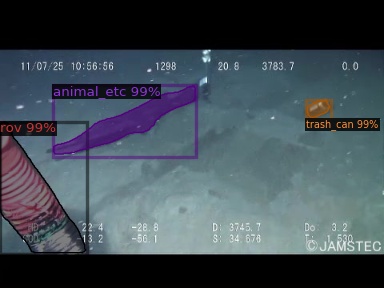} &
    \includegraphics[width=.315\textwidth]{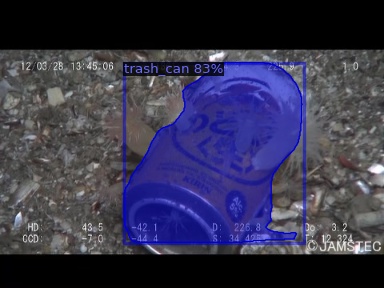} &
    \includegraphics[width=.315\textwidth]{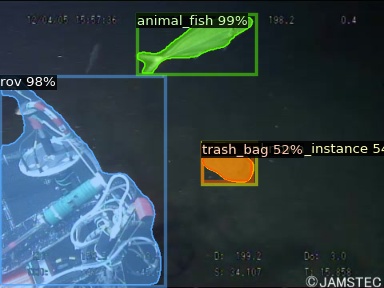} \\
  \raisebox{4.5\normalbaselineskip}[0pt][0pt]{\rotatebox[origin=c]{90}{Seg-Material}} &
    \includegraphics[width=.315\textwidth]{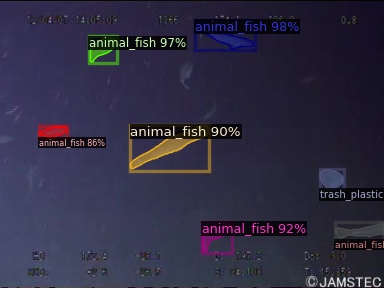} &
    \includegraphics[width=.315\textwidth]{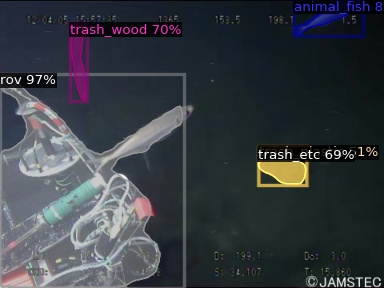} &
    \includegraphics[width=.315\textwidth]{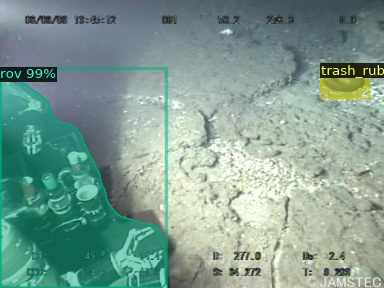} \\
    \raisebox{3.5\normalbaselineskip}[0pt][0pt]{\rotatebox[origin=c]{90}{Bbox-Instance}} &
    \includegraphics[width=.315\textwidth]{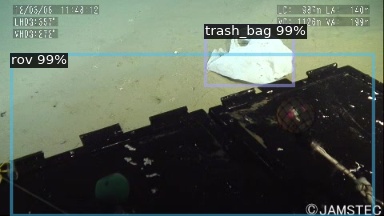} &
    \includegraphics[width=.315\textwidth]{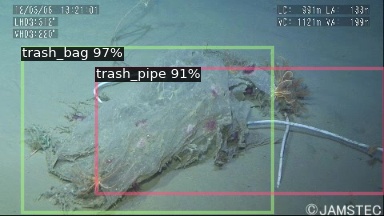} &
    \includegraphics[width=.315\textwidth]{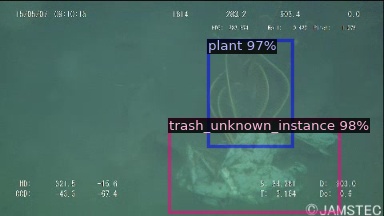} \\
    \raisebox{4.5\normalbaselineskip}[0pt][0pt]{\rotatebox[origin=c]{90}{Bbox-Material}} &
    \includegraphics[width=.315\textwidth]{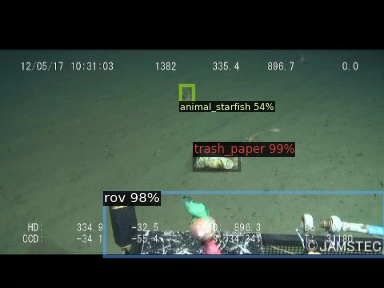} &
    \includegraphics[width=.315\textwidth]{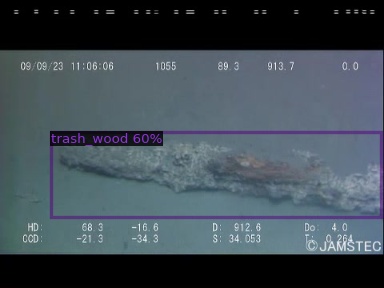} &
    \includegraphics[width=.315\textwidth]{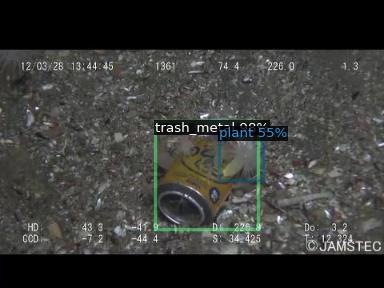} \\
  \end{tabular}
  \caption{Sampled results for object detection and image segmentation for both versions of the TrashCan dataset.}
  \vspace{30mm}
  \label{fig:result_images}
\end{figure*}

Table~\ref{tab:detection_instance_results} shows evaluation metrics of object detection and instance segmentation with each dataset. For both object detection and instance segmentation tasks, the models trained with the instance version dataset achieve higher AP in general. We believe this is because more visually similar objects are grouped into the classes: while most cans look similar, not all metal objects look similar. Fig~\ref{fig:result_images} displays sampled results from object detection and instance segmentation models trained with both versions of the datasets. 
The results include a wide spectrum of object sizes and environments. 
Although the baseline metrics are acceptable considering the challenging nature of this dataset, there is room for improvement in future work, either by increasing the size of the dataset or by employing more advanced models.
It is our hope that the release of this dataset will facilitate further research on this challenging problem, bringing the marine robotics community closer to a solution for the urgent problem of trash detection and removal.

\begin{figure}
    \centering
    \begin{subfigure}[b]{0.75\linewidth}
        \centering
        \includegraphics[width=.99\textwidth]{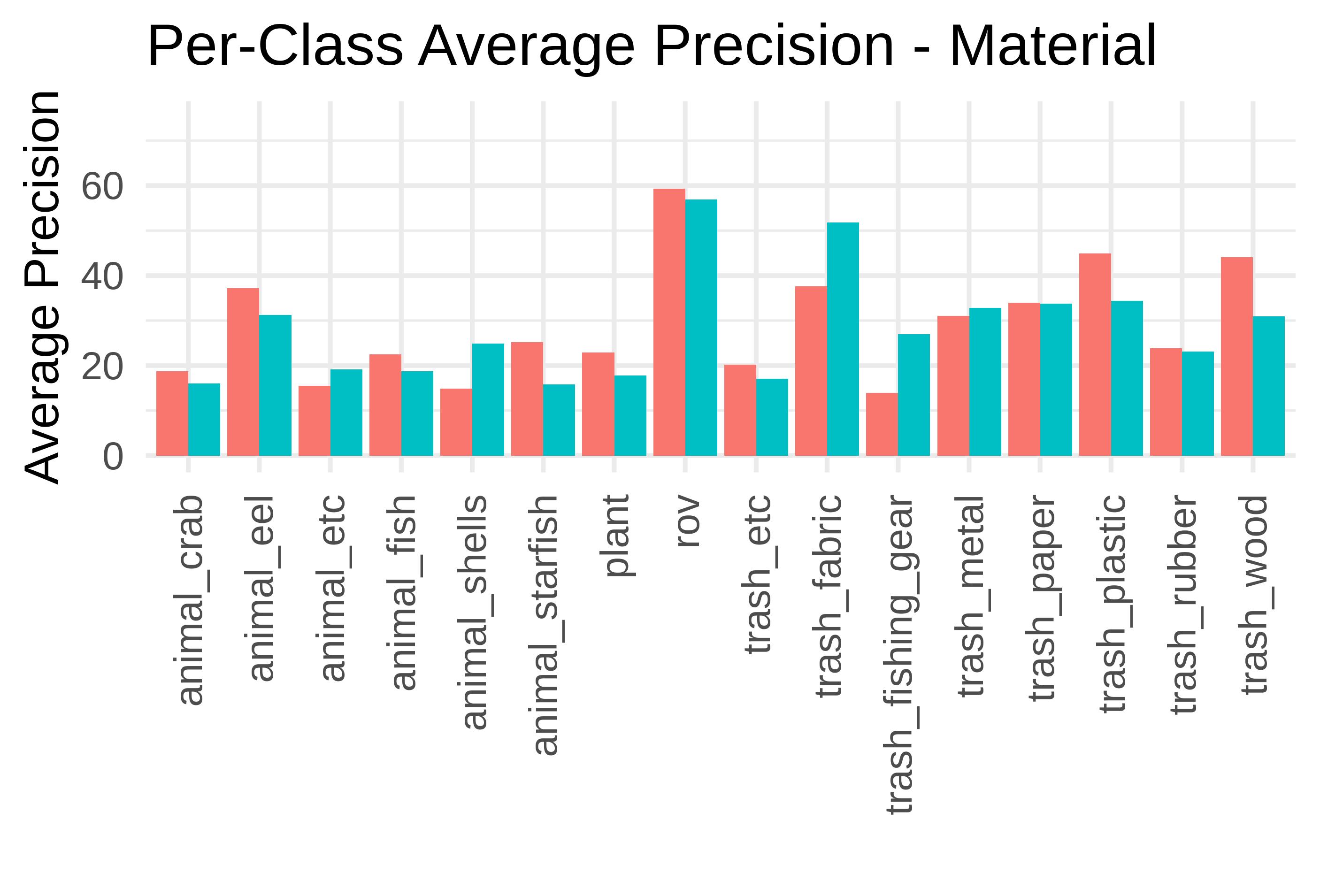}
        \caption{TrashCan-Material}
    \end{subfigure}%
    \hfill
    \begin{subfigure}[b]{0.8\linewidth}
        \centering
        \includegraphics[width=.99\textwidth]{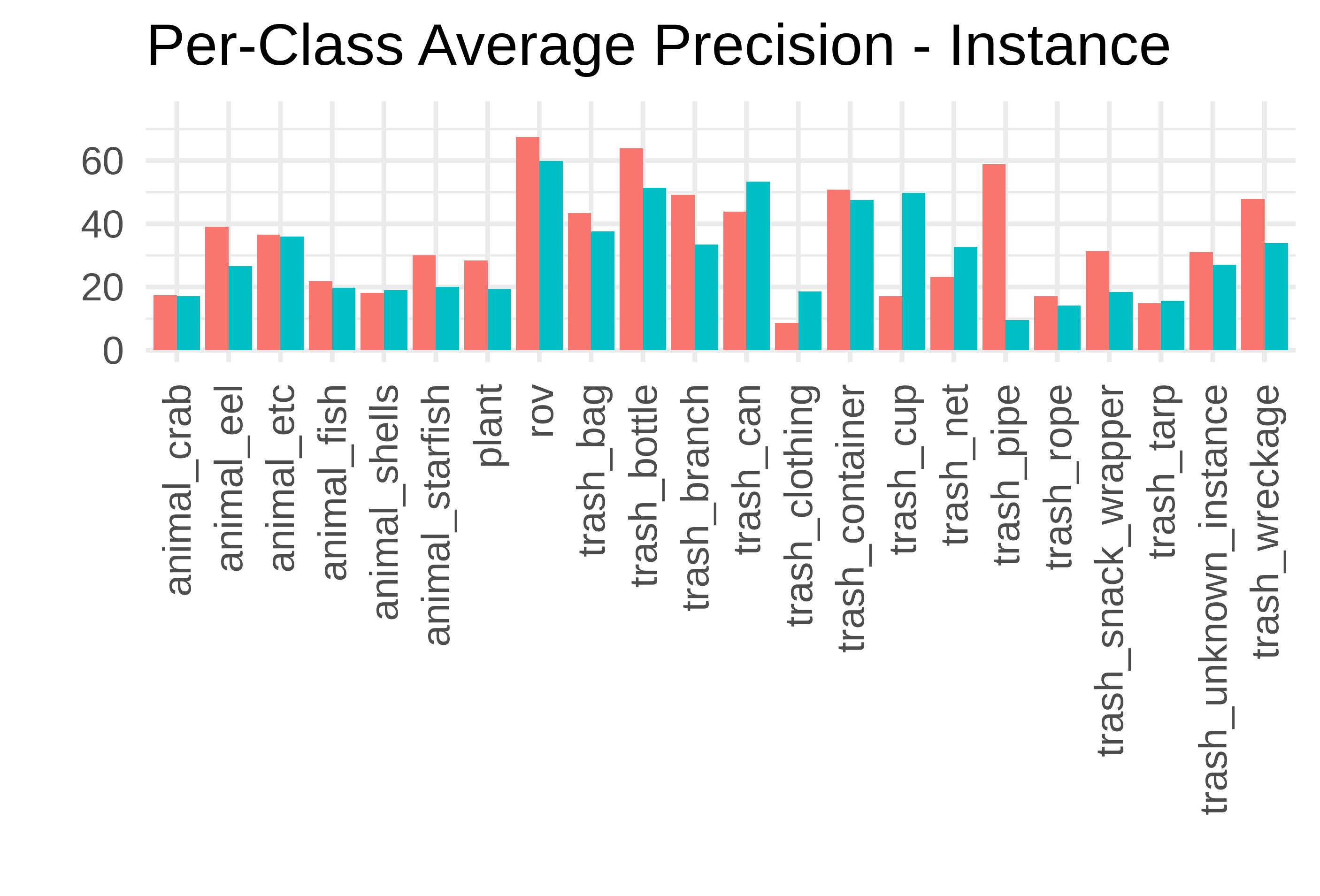}
        \caption{TrashCan-Instance}
    \end{subfigure}%
    \caption{Results from Faster R-CNN (pink) and Mask R-CNN (blue) in terms of per-class average precision.}
    \label{fig:class_ap_results}
\end{figure}


\bibliographystyle{unsrt}
\bibliography{ref}
\end{document}